\documentclass[11pt]{article}

\usepackage[final]{acl}
\usepackage{times}
\usepackage{latexsym}

\usepackage[T1]{fontenc}
\usepackage[utf8]{inputenc}

\usepackage{microtype}

\usepackage{inconsolata}

\usepackage{graphicx}
\usepackage{booktabs}
\usepackage{dblfloatfix}
\usepackage{amsmath}
\usepackage{amssymb}
\usepackage{rotating}
\usepackage{hyperref}
\usepackage{url}

\title{Powerful Training-Free Membership Inference \\Against Fine-Tuned Autoregressive Language Models}

\author{
  David Ilić \quad
  David Stanojević \quad
  Kostadin Cvejoski \\[0.5em]
  JetBrains Research \\
  \texttt{\{david.ilic, david.stanojevic, kostadin.cvejoski\}@jetbrains.com}
}

\begin{document}
\maketitle

\begin{abstract}
Fine-tuned language models pose significant privacy risks, as they may memorize and expose sensitive information from their training data. Membership inference attacks (MIAs) provide a principled framework for auditing these risks, yet existing methods achieve limited detection rates, particularly at the low false-positive thresholds required for practical privacy auditing. We present EZ-MIA, a membership inference attack that exploits a key observation: memorization manifests most strongly at error positions, specifically tokens where the model predicts incorrectly yet still shows elevated probability for training examples. We introduce the Error Zone (EZ) score, which measures the directional imbalance of probability shifts at error positions relative to a pretrained reference model. This principled statistic requires only two forward passes per query and no model training of any kind. On WikiText with GPT-2, EZ-MIA achieves 3.8$\times$ higher detection than the previous state-of-the-art under identical conditions (66.3\% versus 17.5\% true positive rate at 1\% false positive rate), with near-perfect discrimination (AUC 0.98). At the stringent 0.1\% FPR threshold critical for real-world auditing, we achieve 8$\times$ higher detection than prior work (14.0\% versus 1.8\%), requiring no reference model training. These gains extend to larger architectures: on AG News with Llama-2-7B, we achieve 3$\times$ higher detection (46.7\% versus 15.8\% TPR at 1\% FPR). These results establish that privacy risks of fine-tuned language models are substantially greater than previously understood, with implications for both privacy auditing and deployment decisions. Code is available at \url{https://github.com/JetBrains-Research/ez-mia}.
\end{abstract}

\section{Introduction}

The practice of fine-tuning large language models (LLMs) on private datasets has unlocked immense capabilities but also introduces significant privacy risks, as models may memorize and expose sensitive training data \citep{carlini2021extracting}. Membership inference attacks (MIAs), which aim to determine if a specific record was in a model's training set, are the standard tool for auditing these risks \citep{shokri2017membership}. However, existing MIAs are fundamentally limited. Reference-free attacks, which threshold a model's loss or perplexity, suffer from high false-positive rates because they fail to distinguish true memorization from inherently ``easy'' samples \citep{yeom2018privacy}. Reference-based attacks like the Likelihood Ratio Attack (LiRA) mitigate this by calibrating scores against a reference model, but they require unrealistic access to data from the target's training distribution or are computationally prohibitive \citep{carlini2022membership}. Critically, all prior methods reduce a sequence's rich, token-level predictions to a single scalar score, discarding valuable structural information.

Our central insight is that memorization manifests most strongly at \textit{error positions}, specifically tokens where the model fails to predict correctly. At positions where the model succeeds, both the fine-tuned target and pretrained reference typically assign high probability to the correct token, revealing little about membership. But at error positions, a distinctive pattern emerges: for training members, fine-tuning elevates the correct token's probability even when it remains below competing predictions. This residual signal is the signature of memorization that aggregate statistics miss.

We operationalize this insight with EZ-MIA, a membership inference attack of high simplicity and effectiveness. Rather than extracting multiple features or training classifiers, EZ-MIA computes a single statistic: the ratio of upward to downward probability movement at error positions, relative to a pretrained reference model. This Error Zone (EZ) score measures the directional imbalance that memorization induces, a scale-invariant quantity with principled theoretical grounding. The attack requires only two forward passes per query and no model training of any kind: no shadow models, no reference model fine-tuning.

Despite its simplicity, EZ-MIA substantially outperforms prior work across the large majority of dataset and model configurations we evaluate. On WikiText with GPT-2, we achieve 3.8$\times$ higher detection than SPV-MIA \citep{fu2024practical} under identical experimental conditions (66.3\% versus 17.5\% true positive rate at 1\% false positive rate), using only the pretrained base model as reference. At the stringent 0.1\% FPR threshold critical for real-world auditing, we achieve 8$\times$ higher detection than prior work (14.0\% versus 1.8\%), requiring no reference model training. These gains extend to larger architectures: on AG News with Llama-2-7B, we achieve 3$\times$ higher detection (46.7\% versus 15.8\% TPR at 1\% FPR). We further demonstrate that fine-tuning methodology is an important determinant of privacy risk: the same model (GPT-2) on the same data (XSum) yields 82.6\% detection under full fine-tuning but only 1.5\% under LoRA, a 55$\times$ reduction.

These results demonstrate that the privacy risks of fine-tuned language models are substantially greater than previously understood. For privacy auditing, they establish that evaluations using weaker attacks may dramatically underestimate true leakage. For practitioners, they reveal that fine-tuning methodology, not just model scale or training duration, fundamentally shapes privacy risk. EZ-MIA provides a new, more accurate baseline against which both privacy evaluations and defenses must be measured.


\section{Background and Related Work}

Membership inference attacks (MIAs) determine whether a specific data record was used to train a target model, serving both as a direct privacy threat and as a foundation for more sophisticated attacks such as training data extraction \citep{carlini2021extracting}. We review the evolution of these attacks and their application to language models.

\subsection{Membership Inference Foundations}

The study of membership inference began with the shadow model paradigm \citep{shokri2017membership}. This approach trains multiple shadow models to mimic the target's behavior, using their outputs on known members and non-members to train a binary attack classifier. This established the core insight that models behave differently on training versus unseen data. However, shadow model attacks require substantial computational resources and assume the adversary has access to data from a distribution similar to the target's training set.

A simpler approach, the LOSS attack \citep{yeom2018privacy}, observes that training samples typically incur lower loss than non-members, enabling a threshold-based attack. While computationally efficient, this method suffers from high false-positive rates because sample difficulty varies independently of membership: some samples are inherently easy for any model, while others are hard.

\citet{carlini2022membership} formalized membership inference as hypothesis testing and developed the Likelihood Ratio Attack (LiRA), which remains the gold standard for high-precision inference. LiRA calibrates a sample's score using shadow models to factor out its inherent difficulty. Crucially, this work established that attacks should be evaluated by true-positive rate at low false-positive rates (e.g., TPR@0.1\%FPR), a metric critical for practical privacy auditing where high confidence is required. While powerful, LiRA's reliance on training hundreds of shadow models per sample limits its scalability.

\subsection{Attacks and Reference Models}

Modern MIAs can be categorized by their assumptions. \textit{Reference-free attacks} operate solely on the target model's outputs, avoiding the need for auxiliary data. The Neighborhood Attack \citep{mattern2023membership} generates synthetic variants of a query sample and infers membership by comparing the original's likelihood to the average likelihood of its neighbors, at a high computational cost (\textasciitilde 101 forward passes). More efficient methods like MIN-K\% PROB \citep{shi2024detecting} focus on the tokens assigned the lowest probability, intuiting that non-members are more likely to contain such outlier tokens.

\textit{Reference-based attacks} achieve higher precision by calibrating the target's scores against a reference model. The key insight is that a sample with high likelihood under both target and reference models is simply "easy," whereas high likelihood only under the target suggests memorization. The primary challenge for these attacks is the strong assumption that an adversary can obtain data from the target's training distribution to train a suitable reference model.

To address this, \citet{fu2024practical} proposed SPV-MIA, which constructs a reference dataset via self-prompting, prompting the target model itself and fine-tuning a reference model on its generations. While this reduces data assumptions, its membership signal relies on aggregate probabilistic stability across multiple paraphrased inputs, requiring approximately 42 forward passes per sample. Our work departs from prior methods by identifying \textit{where} the membership signal concentrates rather than aggregating across all positions. We show that error positions (where the model predicts incorrectly) carry substantially stronger signal than correct predictions, and exploit this with a principled statistic requiring only two forward passes and no reference model training.

\subsection{Membership Inference Against LLMs}

Applying MIAs to language models presents unique challenges. In foundational work, \citet{carlini2021extracting} demonstrated that LLMs such as GPT-2 memorize and can regurgitate verbatim training data, establishing memorization as a concrete privacy risk.

However, recent work has shown that MIAs against large pretrained models on web-scale data are often ineffective \citep{duan2024membership}. The combination of massive training sets and limited training iterations prevents the strong per-sample memorization that MIAs typically detect. Apparent success in this setting often stems from distribution shifts between member and non-member evaluation sets rather than a genuine membership signal.

The picture is substantially different for fine-tuned models. Fine-tuning uses far smaller datasets, often for multiple epochs, creating a much stronger memorization signal. Studies have consistently shown that fine-tuned models are significantly more vulnerable to membership inference \citep{zhang2025soft, fu2024practical}, though recent work suggests parameter-efficient methods like LoRA may reduce memorization \citep{wang2025leaner}. Our work focuses on this high-risk fine-tuning setting and provides the first quantification of the privacy gap between full fine-tuning and LoRA under a high-precision membership inference attack. Concurrent work has also recognized the insufficiency of aggregate loss for sequence models \citep{rossi2025membership}, but their approach adapts LiRA within the costly shadow model paradigm. In contrast, our work identifies that memorization signal concentrates at error positions and introduces a principled, scale-invariant statistic to measure it, achieving dramatic improvements without any shadow model training.


\section{Methodology}

We present EZ-MIA, a membership inference attack that exploits a key observation: memorization manifests most clearly at positions where the model fails to predict correctly. Our method achieves strong discrimination with only two forward passes and no model training.

\subsection{Threat Model}

We consider a practical threat model where the adversary has:

\begin{enumerate}
    \item \textbf{Query access:} Access to the target model $\theta$, enabling computation of token-level log-probabilities for any input sequence.
    \item \textbf{Reference model access:} Access to the pretrained base model from which the target was fine-tuned, or a public model of comparable architecture.
\end{enumerate}

This threat model aligns with realistic attack scenarios against fine-tuned open-weight models and language model APIs that expose token probabilities. Notably, we require no access to samples from the target model's training distribution and no auxiliary model training.

\subsection{Memorization at Error Positions}

Prior membership inference methods aggregate statistics across all token positions, but this dilutes the membership signal. Our key observation is that \textit{memorization manifests most strongly at positions where the model predicts incorrectly}.

The intuition is straightforward. At positions where the target model's top prediction matches the ground truth, both the fine-tuned target and the pretrained reference typically assign high probability to the correct token; success reveals little about membership. However, at error positions where the model fails to predict correctly, a different pattern emerges: for training members, fine-tuning still elevates the correct token's probability even when it remains below competing tokens. This residual signal (probability mass shifted upward despite prediction failure) is the signature of memorization.

We provide formal and empirical support for this observation in Section \ref{sec:why-error-positions}, and in Appendix~\ref{sec:appendix_ablations} we confirm that restricting to error positions substantially outperforms using success positions.

\subsection{Notation}

Let $\boldsymbol{x} = (x_1, \ldots, x_T)$ be a sequence of tokens. The log-probability assigned by model $\theta$ to token $x_t$ given its prefix is $\ell_\theta^{(t)} = \log p_\theta(x_t \mid \boldsymbol{x}_{<t})$. Given target model $\theta$ and reference model $\hat{\theta}$, we define the token-level log-probability difference:
\begin{equation}
\delta^{(t)} = \ell_\theta^{(t)} - \ell_{\hat{\theta}}^{(t)}
\end{equation}
We define the \textit{error set} $\mathcal{E}$ as positions where the target model's top prediction is incorrect:
\begin{equation}
\mathcal{E} = \{t \mid \arg\max_v p_\theta(v \mid \boldsymbol{x}_{<t}) \neq x_t\}
\end{equation}

\subsection{The Error Zone Score}

Fine-tuning induces probability changes at each error position. We decompose these changes by direction:
\begin{equation}
P = \sum_{t \in \mathcal{E}} [\delta^{(t)}]_+ \qquad N = \sum_{t \in \mathcal{E}} |\,[\delta^{(t)}]_-|
\end{equation}
where $[x]_+ = \max(x, 0)$ and $[x]_- = \min(x, 0)$. Here $P$ represents total probability mass moved \textit{upward} by fine-tuning (relative to the reference), while $N$ represents mass moved \textit{downward}.

The Error Zone score measures the balance of this movement:
\begin{equation}
\text{EZ}(\boldsymbol{x}) = \frac{P}{N}
\end{equation}
This answers a simple question: \textit{of all probability adjustments at error positions, how much more moved up than down?} Memorization creates upward pressure on token probabilities through gradient updates; EZ measures this directional imbalance.

A key property of EZ is scale invariance: multiplying all $\delta^{(t)}$ by a constant $c > 0$ leaves EZ unchanged, since both $P$ and $N$ scale equally. This allows meaningful comparison across sequences with different intrinsic variability; a sequence with volatile predictions and large $|\delta^{(t)}|$ values is compared on equal footing with a predictable sequence showing smaller movements. We provide a principled derivation of EZ, including formal properties and theoretical grounding, in Appendix~\ref{sec:appendix_derivation}.

\subsection{Reference Model}

Our method requires a reference model $\hat{\theta}$ for computing the probability differences $\delta^{(t)}$. We use the pretrained base model checkpoint from before fine-tuning. This choice requires no additional training and provides a natural baseline: it captures general language modeling capabilities without any exposure to the target's training data.

The pretrained reference is both principled and practical. Probability differences $\delta^{(t)}$ directly measure what fine-tuning changed, isolating the effect of training on the target dataset. We evaluate alternative reference constructions in Appendix~\ref{sec:appendix_ablations}.

\subsection{Attack Procedure}

Given a query sequence $\boldsymbol{x}$:

\begin{enumerate}
    \item Compute token-level log-probabilities under the target model $\theta$ and reference model $\hat{\theta}$.
    \item Identify error positions $\mathcal{E}$ where the target's top prediction differs from the ground truth.
    \item Compute the Error Zone score $\text{EZ}(\boldsymbol{x}) = P / N$.
    \item Classify as member if $\text{EZ}(\boldsymbol{x})$ exceeds a threshold $\tau$.
\end{enumerate}

The threshold $\tau$ is chosen to achieve a desired false positive rate. The entire attack requires only two forward passes per query (one through the target model and one through the reference) with no shadow model training, no classifier fitting, and no reference model fine-tuning. This represents an order of magnitude reduction in inference-time computational cost compared to methods like SPV-MIA (\textasciitilde42 forward passes) and the Neighborhood Attack (\textasciitilde101 forward passes), while reducing training-time computational cost to zero.

\section{Experimental Setup}

We evaluate our method across diverse datasets and model architectures to demonstrate its generalizability. This section describes the datasets, target models, baseline methods, and evaluation metrics used in our experiments.

\paragraph{Datasets and Models.} We evaluate on three primary datasets spanning different domains: \textit{AG News} \citep{zhang2015character} (short, topical news), \textit{WikiText-103} \citep{merity2017pointer} (diverse encyclopedic text), and \textit{XSum} \citep{narayan2018dont} (formal journalistic prose). For each, we create disjoint 10k member/10k non-member sets for evaluation, with a separate 500-sequence validation set; we select the checkpoint with lowest validation loss to avoid overfitting artifacts. All sequences are exactly 128 tokens, constructed by concatenating consecutive texts. 

We evaluate across three model scales: \textit{GPT-2} \citep{radford2019language} (124M), \textit{GPT-J} \citep{wang2021gptj} (6B), and \textit{Llama-2} \citep{touvron2023llama2} (7B). GPT-2 is fully fine-tuned for 3 epochs, while larger models use LoRA \citep{hu2022lora} (rank 16, alpha 32) for computational efficiency. To evaluate domain generalization, we additionally test on \textit{Swallow-Code} \citep{fujii2024swallowcode} (Python code) with \textit{Stable-Code-3B} \citep{pinnaparaju2023stablecode}. 

We further evaluate on three additional datasets (Enron emails \citep{metsis2006spam}, PubMed abstracts \citep{cohan2018discourse}, mC4-German \citep{xue2021mt5}) across four additional models spanning 82M to 14B parameters: DistilGPT2 \citep{sanh2019distilbert}, Gemma-3-1B \citep{gemmateam2025gemma3}, DeepSeek-R1-Distill-Llama-8B \citep{deepseek2025r1}, and Qwen3-14B \citep{yang2025qwen3} (Appendix~\ref{sec:extended_eval}). Full training and data processing details are in Appendix \ref{sec:appendix_impl_details}.

\paragraph{Reference Model Construction.} Our attack requires a reference model to compute the membership score. We use the pretrained model checkpoint before any fine-tuning, requiring no additional computation. This provides a natural baseline capturing general language modeling capabilities without exposure to the target's training data.

\paragraph{Baselines.} We compare against representative methods spanning two attack paradigms:

\textit{Reference-free attacks:} LOSS attack \citep{yeom2018privacy}, which thresholds on model loss; Zlib \citep{carlini2021extracting}, which normalizes perplexity by zlib compression length; and Min-K\%++ \citep{zhang2024minkpp}, which aggregates z-score normalized log-probabilities at low-probability tokens.

\textit{Reference-based attacks:} Reference Loss (RefL), which computes the likelihood ratio between the target model and a pretrained reference following the framework of \citet{carlini2022membership}; and SPV-MIA (SPV) \citep{fu2024practical}, the previous state-of-the-art. In tables, we abbreviate Min-K\%++ as MK++.

All baselines are our own implementations evaluated under identical experimental conditions: same models, same fine-tuning protocol, same data splits (10k members, 10k non-members), and same 3-epoch training. For SPV-MIA, we use the official released code executed without modification. This controlled comparison ensures that performance differences reflect the methods themselves rather than experimental setup.

\paragraph{Evaluation Metrics.} We report three complementary metrics: \textit{AUC} (Area Under the ROC Curve) for overall discrimination, \textit{TPR@1\%FPR} (True Positive Rate at 1\% False Positive Rate) for practical auditing scenarios, and \textit{TPR@0.1\%FPR} for high-precision settings where false positives are costly. Following \citet{carlini2022membership}, we focus analysis on the low-FPR metrics, as these determine an attack's practical utility for privacy auditing.

\paragraph{Computational Cost.} Our method requires no training and only two forward passes per query (one through target, one through reference), compared to approximately 42 for SPV-MIA. This efficiency enables practical large-scale auditing.

\paragraph{Computational Resources.}
All experiments were conducted on a single NVIDIA H200 GPU. Fine-tuning and evaluation for each model-dataset configuration completes within one hour.


\section{Results}

\begin{figure*}[t]
    \centering
    \includegraphics[width=\textwidth]{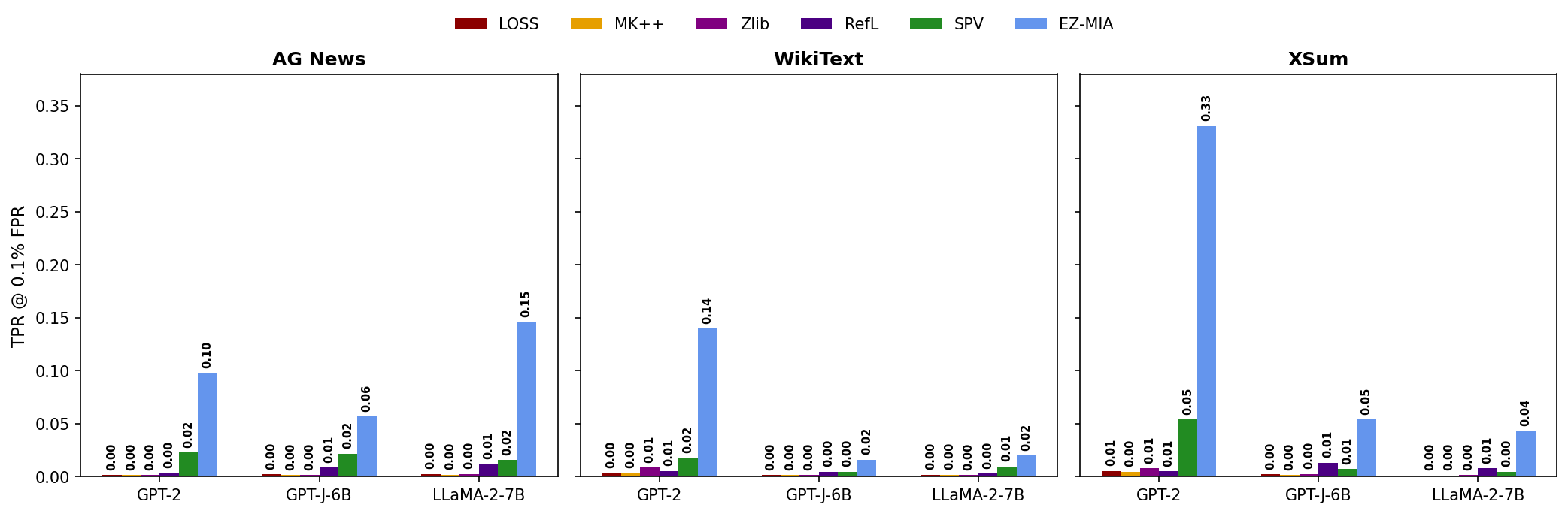}
    \caption{TPR@0.1\%FPR comparison across datasets and models. All methods evaluated under identical experimental conditions. GPT-2 uses full fine-tuning; GPT-J and Llama-2 use LoRA. EZ-MIA achieves up to 9$\times$ higher detection rates than prior work at this stringent threshold critical for privacy auditing.}
    \label{fig:tpr_comparison}
\end{figure*}

Figure~\ref{fig:tpr_comparison} and Table~\ref{tab:main_results} present our main experimental results. We compare EZ-MIA against prior baselines spanning reference-free attacks (LOSS, Zlib, Min-K\%++) and reference-based attacks (Reference Loss, SPV-MIA). Our evaluation encompasses three text domains (AG News, WikiText-103, XSum) and three model scales (GPT-2 124M, GPT-J 6B, Llama-2 7B).

\subsection{Comparison with Prior Methods}

EZ-MIA substantially outperforms all prior attacks in the large majority of configurations. Across our primary evaluation (Table~\ref{tab:main_results}), we achieve average AUC improvements of +0.25 over LOSS, +0.25 over Zlib, and +0.28 over Min-K\%++. The strongest prior attack, SPV-MIA (0.84 average AUC), trails our method by +0.06 AUC on average. Reference Loss, which like our method uses the pretrained model as reference, achieves 0.78 average AUC. These gaps confirm that focusing on error positions captures memorization signals invisible to aggregate statistics.

\subsection{Main Results}

Table~\ref{tab:main_results} presents comprehensive results comparing EZ-MIA against baselines. GPT-2 is fully fine-tuned, while GPT-J and Llama-2 use LoRA (rank 16, alpha 32) for computational efficiency. Under identical experimental conditions, EZ-MIA achieves the highest AUC in the large majority of configurations.

\begin{table*}[!t]
\centering
\small
\resizebox{\textwidth}{!}{%
\begin{tabular}{llcccccccccccc}
\toprule
& & \multicolumn{6}{c}{\textbf{AUC}} & \multicolumn{6}{c}{\textbf{TPR@1\%FPR (\%)}} \\
\cmidrule(lr){3-8} \cmidrule(lr){9-14}
\textbf{Dataset} & \textbf{Model} & \textbf{LOSS} & \textbf{Zlib} & \textbf{MK++} & \textbf{RefL} & \textbf{SPV} & \textbf{EZ-MIA} & \textbf{LOSS} & \textbf{Zlib} & \textbf{MK++} & \textbf{RefL} & \textbf{SPV} & \textbf{EZ-MIA} \\
\midrule
WikiText & GPT-2 & .725 & .735 & .688 & .810 & .899 & \textbf{.984} & 4.1 & 5.3 & 2.7 & 5.2 & 17.5 & \textbf{66.3} \\
WikiText & GPT-J & .563 & .566 & .540 & .615 & .763 & \textbf{.781} & 1.7 & 1.7 & 1.4 & 2.7 & 4.9 & \textbf{10.7} \\
WikiText & Llama-2 & .553 & .555 & .541 & .623 & \textbf{.780} & .771 & 1.7 & 1.7 & 1.3 & 2.5 & 5.4 & \textbf{9.4} \\
AG News & GPT-2 & .742 & .740 & .709 & .773 & .879 & \textbf{.950} & 2.0 & 2.0 & 2.4 & 2.0 & 18.1 & \textbf{39.4} \\
AG News & GPT-J & .704 & .701 & .665 & .828 & .915 & \textbf{.955} & 2.1 & 2.1 & 2.3 & 3.1 & 21.6 & \textbf{42.6} \\
AG News & Llama-2 & .691 & .686 & .639 & .844 & .898 & \textbf{.961} & 1.6 & 1.5 & 1.5 & 3.8 & 15.8 & \textbf{46.7} \\
XSum & GPT-2 & .742 & .743 & .682 & .961 & .927 & \textbf{.993} & 5.5 & 5.6 & 4.2 & 14.7 & 31.4 & \textbf{82.6} \\
XSum & GPT-J & .578 & .577 & .551 & .787 & .761 & \textbf{.860} & 1.8 & 2.0 & 1.5 & 6.2 & 5.6 & \textbf{19.7} \\
XSum & Llama-2 & .579 & .576 & .553 & .799 & .756 & \textbf{.840} & 1.9 & 1.9 & 1.5 & 5.5 & 5.3 & \textbf{17.6} \\
\bottomrule
\end{tabular}
}
\caption{Main results across datasets and models. All baselines are our own implementations under identical conditions. GPT-2 uses full fine-tuning; GPT-J and Llama-2 use LoRA. Bold indicates best per row.}
\label{tab:main_results}
\end{table*}

The results reveal two distinct performance regimes. On fully fine-tuned GPT-2, EZ-MIA achieves strong discrimination: 0.984 AUC on WikiText, 0.950 on AG News, and 0.993 on XSum, with TPR@1\%FPR of 66.3\%, 39.4\%, and 82.6\% respectively. At the stringent 0.1\% FPR threshold, we detect 14.0\%, 9.8\%, and 33.1\% of members on WikiText, AG News, and XSum respectively (Table~\ref{tab:tpr_full}; visualized in Figure~\ref{fig:tpr_comparison}).

Performance on LoRA fine-tuned models (GPT-J, Llama-2) remains strong but substantially lower, with average AUC of 0.86 compared to 0.98 for full fine-tuning. This gap reflects the reduced memorization induced by parameter-efficient fine-tuning, which we analyze in the following subsection. These results generalize to four additional models (82M--14B) across three further domains (Appendix~\ref{sec:extended_eval}).

\subsection{Effect of Fine-tuning Method}

The gap between full fine-tuning and LoRA results in Table~\ref{tab:main_results} conflates fine-tuning method with model architecture. To isolate these factors, we evaluate EZ-MIA on three models under both full fine-tuning and LoRA on XSum, spanning 124M to 7B parameters.

\begin{table*}[!b]
\centering
\begin{tabular}{llccc}
\toprule
\textbf{Model} & \textbf{Fine-tuning} & \textbf{AUC} & \textbf{TPR@1\%} & \textbf{TPR@0.1\%} \\
\midrule
GPT-2 (124M) & Full & 0.993 & 82.6\% & 33.1\% \\
GPT-2 (124M) & LoRA & 0.553 & 1.5\% & 0.1\% \\
\midrule
GPT-2-XL (1.5B) & Full & 0.998 & 98.6\% & 96.1\% \\
GPT-2-XL (1.5B) & LoRA & 0.835 & 7.7\% & 1.1\% \\
\midrule
Llama-2 (7B) & Full & 0.999 & 99.0\% & 98.1\% \\
Llama-2 (7B) & LoRA & 0.840 & 17.6\% & 4.3\% \\
\bottomrule
\end{tabular}
\caption{Effect of fine-tuning method on privacy leakage across scales (XSum, EZ-MIA). Comparing rows within each model isolates fine-tuning method, while comparing across models at the same method isolates scale.}
\label{tab:finetune_method}
\end{table*}

Table~\ref{tab:finetune_method} reveals two clear patterns. First, within each model, full fine-tuning yields dramatically higher leakage than LoRA: TPR@1\%FPR drops by 55$\times$ on GPT-2 (82.6\% $\to$ 1.5\%), 13$\times$ on GPT-2-XL (98.6\% $\to$ 7.7\%), and 6$\times$ on Llama-2 (99.0\% $\to$ 17.6\%). Second, across models at the same fine-tuning method, performance is comparable: all three fully fine-tuned models achieve $\geq$0.993 AUC and $\geq$82.6\% TPR@1\%FPR. This confirms that fine-tuning method, not model scale, is the primary determinant of privacy risk.

The full-vs-LoRA gap narrows at larger scale (55$\times$ at 124M, 6$\times$ at 7B), suggesting that larger LoRA-tuned models still memorize more than smaller ones. Nevertheless, LoRA provides substantial protection at every scale, consistent with recent findings that parameter-efficient fine-tuning reduces memorization \citep{wang2025leaner}. These results imply that privacy audits must account for fine-tuning methodology; evaluations on LoRA-tuned models may dramatically underestimate the risks of full fine-tuning.
Additionally, we find EZ-MIA degrades gracefully as training set size increases from 10k to 250k members (Appendix~\ref{sec:member-scaling}).

\subsection{Computational Efficiency}

EZ-MIA requires only two forward passes per query (one through the target model, one through the reference) and no reference model training. This compares favorably to SPV-MIA, which requires approximately 42 forward passes plus reference model training. The efficiency gain enables practical large-scale privacy auditing with minimal computational overhead.

\subsection{Generalization to Code}

To evaluate domain generalization beyond natural language, we test EZ-MIA on Swallow-Code with Stable-Code-3B (LoRA fine-tuned). We achieve 0.893 AUC and 38.8\% TPR@1\%FPR. This strong performance, comparable to LoRA-tuned natural language models, confirms that EZ-MIA captures domain-agnostic memorization signals and generalizes beyond the text domains.


\section{Analysis}

Beyond comparing against baselines, we analyze when and why EZ-MIA succeeds or fails. We examine how the membership signal evolves during training and characterize the conditions under which our method is most and least effective.

\subsection{Why Error Positions Carry Signal}
\label{sec:why-error-positions}

The results in Section~5 demonstrate that restricting attention to error positions yields substantial gains over aggregate methods. Here we provide a formal explanation grounded in the mechanics of gradient descent, and validate it empirically across models and domains.

Consider a single gradient step on training sequence $x$. The cross-entropy loss at position $t$ is $\mathcal{L}_t = -\ell^{(t)}_\theta = -\log p_\theta(x_t \mid x_{<t})$, and its gradient with respect to the logit vector $z^{(t)}$ takes the form:
\begin{equation}
\frac{\partial \mathcal{L}_t}{\partial z^{(t)}} = p_\theta(\cdot \mid x_{<t}) - e_{x_t}
\end{equation}
where $e_{x_t}$ is the one-hot vector for the ground truth token. The gradient update therefore pushes the logits in the direction $e_{x_t} - p_\theta(\cdot \mid x_{<t})$, with the component corresponding to the correct token equal to $1 - p_\theta(x_t \mid x_{<t})$.

At success positions, where the model already predicts $x_t$ correctly, $p_\theta(x_t \mid x_{<t})$ is large. The residual $1 - p_\theta(x_t \mid x_{<t})$ is correspondingly small, and fine-tuning produces little change. Both members and non-members look similar at these positions, because the model's confidence leaves little room for the gradient to act.

At error positions, the picture is different. Here $p_\theta(x_t \mid x_{<t})$ is small, as the model has placed its probability mass elsewhere. The residual is close to 1, producing a large gradient that pushes the correct token's probability upward. Crucially, this push occurs \emph{only for training members}: non-members receive no direct gradient signal, and any probability changes at their error positions reflect indirect generalization from other training examples, which lacks the strong, consistent upward pressure that direct training on the sequence produces. This asymmetry is the structural reason that memorization concentrates at error positions.

We test this prediction directly. For each of three configurations spanning different model scales, domains, and fine-tuning methods, we compute rank-improvement frequency at every token position across 10,000 members and 10,000 non-members. A position counts as rank-improved if fine-tuning lowered the rank of the correct token relative to the reference model: $\text{rank}_\theta(x_t \mid x_{<t}) < \text{rank}_{\hat{\theta}}(x_t \mid x_{<t})$. We partition positions into error and success and report the frequency of rank improvement for each group.

\begin{table*}[t]
\centering
\small
\begin{tabular}{l cc c cc c}
\toprule
& \multicolumn{3}{c}{\textbf{Error Positions}} & \multicolumn{3}{c}{\textbf{Success Positions}} \\
\cmidrule(lr){2-4} \cmidrule(lr){5-7}
\textbf{Configuration} & \textbf{Members} & \textbf{Non-mem.} & \textbf{Gap} & \textbf{Members} & \textbf{Non-mem.} & \textbf{Gap} \\
\midrule
WikiText / GPT-2 (124M, full FT) & .691 & .518 & \textbf{.173} & .321 & .278 & .043 \\
AG News / Llama-2-7B (LoRA) & .564 & .477 & \textbf{.087} & .151 & .131 & .020 \\
Swallow / Stable-Code-3B (LoRA) & .495 & .404 & \textbf{.091} & .077 & .063 & .014 \\
\bottomrule
\end{tabular}
\caption{Rank-improvement frequency at error vs.\ success positions. Fine-tuning improves correct-token rank far more often for members at error positions, with gaps 4--6$\times$ larger than at success positions (all $p < 0.0001$, Mann-Whitney $U$).}
\label{tab:rank-improvement}
\end{table*}

The results in Table \ref{tab:rank-improvement} confirm the gradient analysis. At error positions, members show rank improvement 50--69\% of the time compared to 40--52\% for non-members, producing gaps of 0.087--0.173. At success positions, the gaps shrink to 0.014--0.043: 4--6$\times$ smaller. This pattern holds across GPT-2 (124M, full fine-tuning), Llama-2-7B (LoRA), and Stable-Code-3B (LoRA), spanning encyclopedic text, news, and code. The concentration of membership signal at error positions is not an artifact of a particular model or domain, but a structural consequence of how gradient descent distributes its updates.

\subsection{Relationship to Training Dynamics}

The analysis in this and the following subsection uses XSum with fully fine-tuned GPT-2-XL (1.5B parameters) and 1k members/1k non-members, chosen because the larger model exhibits stronger memorization, making failure modes more visible. We begin by investigating how EZ-MIA performance evolves as fine-tuning progresses, evaluating at checkpoints after each training epoch.

\begin{figure}[ht]
    \centering
    \includegraphics[width=0.85\columnwidth]{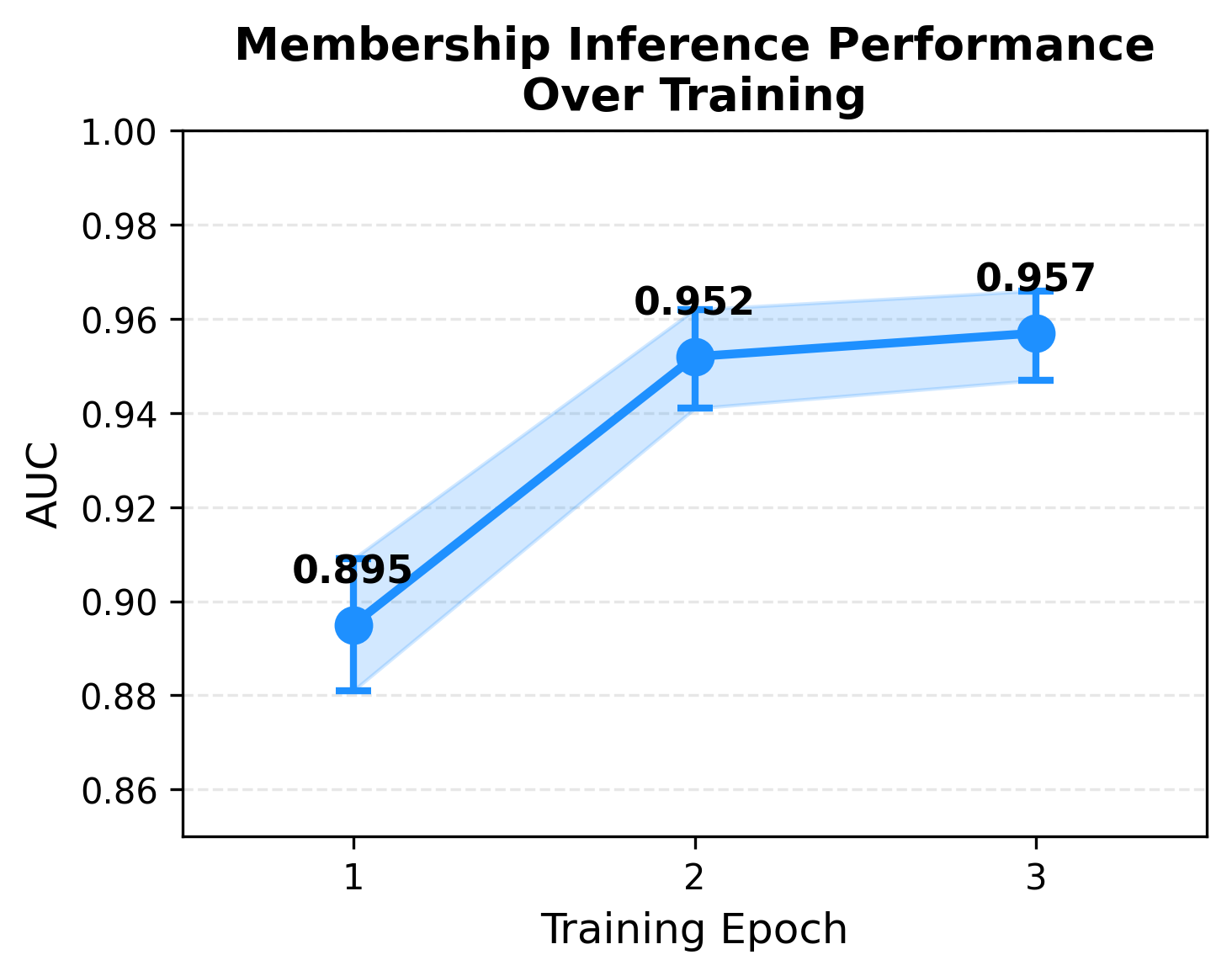}
    \caption{EZ-MIA performance over training epochs on XSum/GPT-2-XL. Privacy leakage emerges early (epoch 1: AUC = 0.895) and increases with continued training. Error bars show 95\% bootstrap confidence intervals (1,000 resamples of evaluation sequences).}
    \label{fig:training_dynamics}
\end{figure}

Figure~\ref{fig:training_dynamics} shows that AUC increases monotonically from 0.895 after epoch 1 to 0.957 after epoch 3. Notably, even a single epoch of fine-tuning induces sufficient memorization for strong membership inference (TPR@1\%FPR = 54.1\% at epoch 1, rising to 90.2\% at epoch 3). This indicates that privacy risks emerge early in training.

More striking is the relationship between EZ-MIA performance and overfitting. We compute the train-test loss gap (a standard measure of overfitting) at each checkpoint and observe that AUC increases monotonically with this gap across all three epochs. This pattern is consistent with our theoretical framework: the EZ score measures precisely the probability boost that fine-tuning confers on training examples, which is the token-level manifestation of overfitting. While three checkpoints are insufficient for formal statistical inference, this connection suggests that EZ-MIA could serve not only as a post-hoc auditing tool but also as a real-time privacy monitor during training, a hypothesis that warrants further investigation with more fine-grained training dynamics.

\subsection{Robustness and Failure Modes}

We examine whether EZ-MIA performance depends on sample difficulty and characterize when the method fails.

\paragraph{Robustness to Sample Difficulty.} One concern is that EZ-MIA might only succeed on ``easy'' samples while failing on difficult ones. We partition sequences into quartiles by reference model perplexity (a proxy for difficulty) and compute AUC within each quartile on XSum/GPT-2-XL:

\begin{center}
\begin{tabular}{lcccc}
\toprule
& Q1 (easy) & Q2 & Q3 & Q4 (hard) \\
\midrule
AUC & 0.924 & 0.893 & 0.878 & 0.887 \\
\bottomrule
\end{tabular}
\end{center}

Performance is consistent across difficulty levels (range: 0.046), with no systematic degradation on hard samples. EZ-MIA's focus on error positions may explain this robustness: difficult samples have more error positions, providing more signal rather than less.

\paragraph{Failure Mode Characterization.} At 10\% FPR on XSum/GPT-2-XL, we examine false negatives (members incorrectly classified as non-members) and false positives (non-members incorrectly classified as members):

\begin{table}[h]
\centering
\small
\begin{tabular}{lccc}
\toprule
& \textbf{Avg Tokens} & \textbf{Avg Errors} & \textbf{Mean $\boldsymbol{\delta}$} \\
\midrule
All Members & 354 & 173 & $+0.30$ \\
False Negatives & 430 & 230 & $-0.15$ \\
\midrule
All Non-members & 363 & 189 & $+0.02$ \\
False Positives & 94 & 46 & $+0.41$ \\
\bottomrule
\end{tabular}
\caption{Characteristics of failure cases on XSum/GPT-2-XL evaluated on 1k members and 1k nonmembers, with unrestricted sequence length. False negatives are longer sequences with more errors and negative mean $\delta$; false positives are shorter sequences where the reference model performs unusually poorly.}
\label{tab:failure_modes}
\end{table}

False negatives (missed members) are substantially longer than average (430 vs.\ 354 tokens) with more error positions (230 vs.\ 173) and negative mean $\delta$; these are members on which the model actually performs \textit{worse} than the reference, perhaps due to distribution shift within the training set. False positives are notably shorter (94 vs.\ 363 tokens) with fewer errors (46 vs.\ 189) and high mean $\delta$; these are non-members where the reference model happens to perform poorly, creating a spurious signal.

These failure modes are predictable from sequence characteristics rather than random. Practitioners should exercise additional caution when auditing sequences of extreme relative length, where EZ-MIA's reliability degrades.


\section{Conclusion}

We have presented EZ-MIA, a membership inference attack against fine-tuned language models that achieves dramatic improvements over prior work through a simple insight: memorization manifests most strongly at error positions. Rather than aggregating statistics across all tokens, we focus on positions where the model predicts incorrectly, measuring the directional imbalance of probability shifts relative to a pretrained reference. This Error Zone score captures memorization signal that previous methods miss entirely.

The magnitude of our improvements warrants emphasis. At false positive rates relevant for practical privacy auditing, we detect members at rates up to 9$\times$ higher than prior methods. On fully fine-tuned models, we achieve 83\% TPR at 1\% FPR on XSum with GPT-2. Yet our method requires only two forward passes per query and no model training of any kind, an order of magnitude more efficient than prior reference-based attacks.

Our results carry significant implications. For privacy auditing, they demonstrate that current evaluation practices using weaker attacks may substantially underestimate true privacy risks. 

Beyond auditing, EZ-MIA has direct implications for training data extraction. Modern extraction pipelines \citep{carlini2021extracting} use a two-stage approach: generating candidate sequences, then filtering with a membership inference attack to identify genuine training data. EZ-MIA's strong performance at low false-positive rates, which is the regime that determines filtering precision, suggests it could substantially improve extraction recall as a drop-in replacement for the filtering stage.

For practitioners, our results reveal that fine-tuning methodology fundamentally shapes privacy exposure: the same model yields 83\% detection under full fine-tuning but only 1.5\% under LoRA, a 55$\times$ reduction. For defense design, EZ-MIA establishes a new baseline against which mitigations should be evaluated. 

Ultimately, this work demonstrates that the privacy risks of fine-tuned models are greater than previously understood, underscoring the need for more rigorous privacy evaluation and methodology-aware deployment decisions.

\section*{Acknowledgements}

We thank Evgeny Grigorenko and Egor Bogomolov for feedback and infrastructure support. This work was supported by JetBrains Research.

\section*{Limitations}

Our method has several important limitations.

First, while EZ-MIA outperforms prior work in the large majority of configurations, it is not universally superior. On WikiText with Llama-2-7B, SPV-MIA achieves marginally higher AUC (0.780 vs 0.771).

Second, our evaluation is scoped to the fine-tuning setting. Applying EZ-MIA to models pretrained on web-scale data remains an open challenge, as the memorization signal is far weaker and more diffuse in that regime.

Third, while we demonstrate that fine-tuning method dramatically affects privacy risk, our LoRA experiments use a single configuration (rank 16, alpha 32). The relationship between LoRA hyperparameters and privacy leakage as measured by EZ-MIA remains unexplored; different rank or alpha values may yield different vulnerability profiles.

Finally, the EZ score requires identifying error positions, which assumes access to ground truth tokens. This assumption holds for membership inference and data extraction (where the attacker queries with candidate training sequences) but may limit applicability to other privacy attacks.

\section*{Ethics Statement}

The development of more powerful membership inference attacks presents a clear dual-use concern. The techniques presented in this paper could, in principle, be used by malicious actors to probe deployed language models and infer the presence of sensitive information in their training sets, potentially deanonymizing individuals or revealing confidential data. We acknowledge this risk and have carefully considered the ethical implications of our work.

Our primary motivation is defensive. The field of AI privacy operates on the principle that one cannot defend against a threat that is not well understood. Existing MIA benchmarks, by underestimating the true potential of these attacks, may provide a false sense of security. Our work serves as a more accurate ``yardstick'' for privacy risk, demonstrating that leakage from fine-tuned models is far more severe than previously established.

The simplicity of EZ-MIA amplifies both its risks and its benefits. Requiring only two forward passes and no model training, it is accessible to a wider range of actors, but this same accessibility enables broader adoption for legitimate privacy auditing. Organizations with limited computational resources can now conduct rigorous privacy evaluations that were previously impractical.

Our finding that fine-tuning methodology dramatically affects privacy risk carries immediate practical value. Practitioners can make informed decisions about the privacy-utility tradeoffs of full fine-tuning versus parameter-efficient methods. This actionable guidance (that LoRA reduces membership inference vulnerability by an order of magnitude) may prevent privacy harms that would otherwise occur.

By providing a stronger, more efficient auditing tool and releasing our code to the research community, we aim to empower developers, researchers, and regulators to:
\begin{itemize}
    \item Conduct more realistic and rigorous privacy audits before deployment.
    \item Develop and calibrate stronger privacy-preserving defenses against a realistic threat model.
    \item Make informed choices about fine-tuning methodology based on privacy requirements.
    \item Foster greater transparency and accountability regarding the privacy properties of deployed AI systems.
\end{itemize}

We believe that the transparent, rigorous quantification of privacy risks is an essential prerequisite for developing technology that is both powerful and safe.

\bibliography{custom}

\appendix

\section{Implementation Details}
\label{sec:appendix_impl_details}

\subsection{Data Processing}

For each dataset, we sample sequences from the training split to create two disjoint evaluation sets: (1) 10,000 target members used to fine-tune the target model and labeled as members for evaluation, and (2) 10,000 target non-members labeled as non-members for evaluation. We maintain a held-out validation set of 500 sequences for model selection during fine-tuning. All sequences are exactly 128 tokens, constructed by concatenating consecutive texts and discarding overflow.

\subsection{Target Model Fine-tuning}

Table~\ref{tab:finetune_config} summarizes the fine-tuning configuration for each model. All models use the AdamW optimizer \citep{loshchilov2019decoupled}. We select the checkpoint with the lowest validation loss to mitigate overfitting artifacts that could confound membership signals.

\begin{table}[h]
\centering
\small
\resizebox{\columnwidth}{!}{%
\begin{tabular}{llcccc}
\toprule
\textbf{Model} & \textbf{Method} & \textbf{Epochs} & \textbf{LR} & \textbf{Batch} & \textbf{LoRA Config} \\
\midrule
GPT-2 (124M) & Full & 3 & $1 \times 10^{-4}$ & 16 & — \\
GPT-2-XL (1.5B) & Full & 3 & $1 \times 10^{-4}$ & 16 & — \\
GPT-2-XL (1.5B) & LoRA & 3 & $1 \times 10^{-4}$ & 16 & $r{=}16$, $\alpha{=}32$ \\
GPT-J (6B) & LoRA & 3 & $1 \times 10^{-4}$ & 16 & $r{=}16$, $\alpha{=}32$ \\
Llama-2 (7B) & LoRA & 3 & $1 \times 10^{-4}$ & 16 & $r{=}16$, $\alpha{=}32$ \\
Stable-Code (3B) & LoRA & 3 & $1 \times 10^{-4}$ & 16 & $r{=}16$, $\alpha{=}32$ \\
Llama-2 (7B) & Full & 3 & $1 \times 10^{-4}$ & 16 & — \\
DistilGPT2 (82M) & Full & 3 & $1 \times 10^{-4}$ & 16 & — \\
Gemma-3-1B (1B) & LoRA & 3 & $1 \times 10^{-4}$ & 16 & $r{=}16$, $\alpha{=}32$ \\
DeepSeek-8B (8B) & LoRA & 3 & $1 \times 10^{-4}$ & 16 & $r{=}16$, $\alpha{=}32$ \\
Qwen3-14B (14B) & LoRA & 3 & $1 \times 10^{-4}$ & 16 & $r{=}16$, $\alpha{=}32$ \\
Qwen-2.5-1.5B (1.5B) & LoRA & 3 & $1 \times 10^{-4}$ & 16 & $r{=}16$, $\alpha{=}32$ \\
\bottomrule
\end{tabular}%
}
\caption{Fine-tuning configuration for target models. LoRA configurations use dropout 0.05.}
\label{tab:finetune_config}
\end{table}


\section{Baseline Implementation Details}
\label{sec:appendix_baselines}

All baseline attacks are evaluated under identical experimental conditions to EZ-MIA: same models, same fine-tuning protocol (3 epochs with validation-based checkpoint selection), same data splits (10k members, 10k non-members), and same evaluation metrics.

\paragraph{Reference-Free Attacks.} We implement LOSS \citep{yeom2018privacy}, Zlib \citep{carlini2021extracting}, and Min-K\%++ \citep{zhang2024minkpp} following their original descriptions. Our implementations are included in the repository.

\paragraph{Reference Loss.} We implement the likelihood ratio between target and pretrained reference models following the framework of \citet{carlini2022membership}. The score is computed as $L_{\text{ref}} - L_{\text{target}}$, where $L$ denotes mean negative log-likelihood.

\paragraph{SPV-MIA.} For SPV-MIA \citep{fu2024practical}, we use the official released code executed without modification. Since the paper reports multiple algorithm variants, we use Algorithm 2 (the best-performing version with publicly available code). The code is included in our repository.

This controlled experimental setup ensures that performance differences between methods reflect the attacks themselves rather than experimental conditions.


\section{TPR at Stringent Thresholds}
\label{sec:appendix_tpr}

Table~\ref{tab:tpr_full} presents TPR@0.1\%FPR results across all configurations. This stringent threshold is critical for privacy auditing applications where false positives are costly.

\begin{table}[t]
\centering
\small
\setlength{\tabcolsep}{4pt}
\begin{tabular}{llrrr}
\toprule
\textbf{Dataset} & \textbf{Model} & \textbf{EZ-MIA} & \textbf{SPV} & \textbf{Mult.} \\
\midrule
WikiText & GPT-2 & 14.0\% & 1.75\% & 8.0$\times$ \\
WikiText & GPT-J & 1.6\% & 0.47\% & 3.4$\times$ \\
WikiText & Llama-2 & 2.0\% & 0.97\% & 2.1$\times$ \\
AG News & GPT-2 & 9.8\% & 2.29\% & 4.3$\times$ \\
AG News & GPT-J & 5.7\% & 2.18\% & 2.6$\times$ \\
AG News & Llama-2 & 14.6\% & 1.57\% & 9.3$\times$ \\
XSum & GPT-2 & 33.1\% & 5.41\% & 6.1$\times$ \\
XSum & GPT-J & 5.4\% & 0.73\% & 7.4$\times$ \\
XSum & Llama-2 & 4.3\% & 0.49\% & 8.8$\times$ \\
\midrule
\multicolumn{2}{l}{\textbf{Average}} & \textbf{10.1\%} & 1.76\% & \textbf{5.7$\times$} \\
\bottomrule
\end{tabular}
\caption{TPR@0.1\%FPR comparison. EZ-MIA achieves higher detection rates at this stringent threshold across all configurations, with improvements ranging from 2$\times$ to 9$\times$.}
\label{tab:tpr_full}
\end{table}

\section{Principled Derivation of the Error Zone Score}
\label{sec:appendix_derivation}

This appendix provides a theoretical grounding for the Error Zone score, deriving its form from first principles and validating the underlying assumptions empirically.

\subsection{The Detection Problem}

Let $\delta_t = \log p_\theta(x_t \mid x_{<t}) - \log p_{\hat{\theta}}(x_t \mid x_{<t})$ denote the log-probability shift at position $t$ after fine-tuning, where $\theta$ is the fine-tuned target and $\hat{\theta}$ is the pretrained reference. We restrict attention to error positions $\mathcal{E}$ where the fine-tuned model does not predict the correct token.

\textbf{Goal:} From the vector $\boldsymbol{\delta} = (\delta_t)_{t \in \mathcal{E}}$, determine whether sequence $\boldsymbol{x}$ was in the training set.

\subsection{The Memorization Signal}

We posit an additive model for the log-probability shift:
\begin{equation}
\delta_t = M_t + G_t
\end{equation}
where:
\begin{itemize}
    \item $M_t \geq 0$ is the memorization effect: the direct consequence of gradient updates on $\boldsymbol{x}$
    \item $G_t$ is the generalization effect: the consequence of training on other sequences
\end{itemize}

The key assumption is that memorization is non-negative. When gradient descent optimizes $-\log p_\theta(x_t \mid x_{<t})$, it increases $p_\theta(x_t \mid x_{<t})$. Training on a sequence cannot systematically decrease the probability of its own tokens.

For non-members, $M_t = 0$ by definition. Thus:
\begin{center}
\begin{tabular}{lc}
\toprule
& Expected $\delta_t$ \\
\midrule
Non-member & $G_t$ \\
Member & $M_t + G_t \geq G_t$ \\
\bottomrule
\end{tabular}
\end{center}

The signal we seek is the \textit{upward shift} induced by $M_t \geq 0$.

\subsection{Why Raw Sums Fail}

The natural first attempt is $\sum_t \delta_t$. If memorization adds positive mass, members should have larger sums. But this conflates two distinct quantities:

\begin{enumerate}
    \item \textbf{Signal:} The memorization contribution $\sum_t M_t$
    \item \textbf{Scale:} The intrinsic variability of $\boldsymbol{\delta}$ for sequence $\boldsymbol{x}$
\end{enumerate}

Different sequences exhibit wildly different scales. A sequence with rare or unpredictable tokens will have large $|\delta_t|$ values regardless of membership; the model's probability estimates swing more dramatically on harder tokens.

\begin{center}
\small
\setlength{\tabcolsep}{4pt}
\begin{tabular}{lccc}
\toprule
Sequence & $\boldsymbol{\delta}$ & $\sum \delta_t$ & Pattern \\
\midrule
A (predictable) & $(0.1, 0.2, -0.1)$ & $0.2$ & 3/4 up \\
B (volatile) & $(1.0, 2.0, -1.0)$ & $2.0$ & 3/4 up \\
\bottomrule
\end{tabular}
\end{center}

Sequence B has a 10$\times$ larger sum, but both have the same pattern: three-quarters of the movement is upward. Comparing raw sums across sequences conflates the membership signal with sequence-specific volatility.

\subsection{Decomposing Probability Movement}

We reframe the problem in terms of probability mass movement. Fine-tuning induces changes at each error position with a direction (up or down) and magnitude. Define:
\begin{equation}
P = \sum_{t \in \mathcal{E}} [\delta_t]_+ \qquad N = \sum_{t \in \mathcal{E}} |[\delta_t]_-|
\end{equation}
where $[x]_+ = \max(x, 0)$ and $[x]_- = \min(x, 0)$.

\begin{itemize}
    \item $P$ = total probability mass moved upward
    \item $N$ = total probability mass moved downward
    \item $P + N$ = total movement (the ``budget'' of change)
\end{itemize}

This decomposition is exhaustive: every probability adjustment is either up or down.

\subsection{The Ratio Formulation}

Rather than asking ``how much did probabilities increase?'' (scale-dependent), we ask:

\begin{quote}
\textit{Of all probability movement that occurred, what fraction was upward?}
\end{quote}

This fraction is $f_{\text{up}} = P / (P + N)$, which is scale-invariant: multiplying all $\delta_t$ by a constant $c > 0$ leaves $f_{\text{up}}$ unchanged.

The Error Zone score uses the equivalent odds formulation:
\begin{equation}
\text{EZ} = \frac{P}{N}
\end{equation}
The relationship to the fraction form is monotonic: $f_{\text{up}} = \text{EZ} / (1 + \text{EZ})$.

\textbf{Why odds?} The odds formulation offers several advantages:
\begin{enumerate}
    \item Interpretability: $\text{EZ} = 3$ means ``three units of probability moved up for every one that moved down.''
    \item Multiplicative structure: Effects that scale the imbalance act multiplicatively on EZ.
    \item Unbounded range: $\text{EZ} \in (0, \infty)$ with $\text{EZ} = 1$ as the neutral point, avoiding ceiling effects when the signal is strong.
\end{enumerate}

\subsection{Why EZ Captures Memorization}

Under our additive model, what happens when memorization is present? For a member, $\delta_t = M_t + G_t$ with $M_t \geq 0$. Compared to a non-member (where $\delta_t = G_t$), adding $M_t > 0$ has the following effects:

\begin{center}
\small
\setlength{\tabcolsep}{3pt}
\begin{tabular}{lll}
\toprule
Original $G_t$ & Effect of $M_t > 0$ & Impact \\
\midrule
$G_t > 0$ & $\delta_t$ increases & $P \uparrow$ \\
$G_t < 0$, $|G_t| > M_t$ & $\delta_t \to 0$ & $N \downarrow$ \\
$G_t < 0$, $|G_t| < M_t$ & $\delta_t$ flips positive & $P \uparrow, N \downarrow$ \\
\bottomrule
\end{tabular}
\end{center}

In all cases, the ratio $P/N$ increases (or stays the same if $M_t = 0$). The signature of memorization is an elevated EZ.

\subsection{Empirical Validation of Assumptions}

\paragraph{Non-negative memorization ($\mathbb{E}[M_t] \geq 0$).}

We test our core assumption that memorization induces non-negative probability shifts by partitioning sequences into difficulty bins based on reference model perplexity and computing the mean $\delta$ at error positions for members versus non-members. If this assumption holds, the member mean should exceed the non-member mean across all difficulty levels.

\begin{center}
\small
\setlength{\tabcolsep}{3pt}
\begin{tabular}{lccc}
\toprule
Bin & $\mathbb{E}[\delta]$ Mem. & $\mathbb{E}[\delta]$ Non-mem. & $p$ \\
\midrule
Q1 (easy) & 0.312 & 0.041 & $< 0.0001$ \\
Q2 & 0.298 & 0.029 & $< 0.0001$ \\
Q3 & 0.291 & 0.018 & $< 0.0001$ \\
Q4 & 0.287 & 0.012 & $< 0.0001$ \\
Q5 (hard) & 0.279 & 0.005 & $< 0.0001$ \\
\bottomrule
\end{tabular}
\end{center}

\textbf{Finding:} $\mathbb{E}[\delta_{\text{member}}] > \mathbb{E}[\delta_{\text{non-member}}]$ in all five difficulty bins with $p < 0.0001$. The non-negative memorization assumption is strongly supported.

\paragraph{Distributional structure.}

A stronger assumption would be that member and non-member $\delta$ distributions differ only by a location shift. We test this by removing the mean difference and comparing distributions via the Kolmogorov-Smirnov test.

After mean-shift removal, we observe KS statistic $= 0.120$ ($p < 0.0001$), indicating the distributions are not identical. However, examining quantiles reveals that the distributions align closely across the central 80\% of probability mass:

\begin{center}
\small
\setlength{\tabcolsep}{4pt}
\begin{tabular}{lccc}
\toprule
Quantile & Shifted Mem. & Non-mem. & Diff. \\
\midrule
10th & $-0.599$ & $-0.518$ & $-0.081$ \\
25th & $-0.358$ & $-0.247$ & $-0.111$ \\
50th & $-0.115$ & $-0.021$ & $-0.094$ \\
75th & $+0.223$ & $+0.205$ & $+0.018$ \\
90th & $+0.699$ & $+0.493$ & $+0.206$ \\
\bottomrule
\end{tabular}
\end{center}

The divergence concentrates in the right tail (90th percentile and above), where members show excess positive mass. While the violation is statistically significant, we judge the effect size as modest and not practically significant.


\section{Ablations and Design Choices}
\label{sec:appendix_ablations}

This appendix provides empirical justification for key design decisions in EZ-MIA. All experiments in this section use a preliminary evaluation with 1,000 members and 1,000 non-members on XSum with GPT-2-XL. While absolute performance differs from the full 10k/10k evaluation in the main paper, relative comparisons between design choices remain valid.

\subsection{Why Error Positions?}

\paragraph{Error vs.\ Success Positions.}

We compare EZ computed at error positions (where target's top prediction $\neq$ ground truth) versus success positions (where top prediction $=$ ground truth).

\begin{center}
\begin{tabular}{lc}
\toprule
Position Type & AUC \\
\midrule
Error positions & 0.895 \\
Success positions & 0.797 \\
\bottomrule
\end{tabular}
\end{center}

Restricting to error positions provides a +0.1 AUC improvement. This confirms our key insight: the membership signal concentrates where the model fails.

\paragraph{Error Count Distribution.}

Members exhibit fewer errors than non-members on average:

\begin{center}
\begin{tabular}{lcc}
\toprule
Group & Mean Errors & Std \\
\midrule
Members & 172.97 & 76.65 \\
Non-members & 189.31 & 81.91 \\
\bottomrule
\end{tabular}
\end{center}

This difference is significant ($p < 0.0001$) but the correlation with membership is weak ($r = -0.102$), confirming that EZ captures signal beyond simple error counting.

\subsection{Reference Model Variants}

\paragraph{Reference Model Quality.}

We compare different reference model choices:

\begin{center}
\begin{tabular}{lc}
\toprule
Reference Model & AUC \\
\midrule
Pretrained GPT-2-XL & 0.895 \\
Random initialization & 0.591 \\
Self-reference (target as reference) & 0.500 \\
\bottomrule
\end{tabular}
\end{center}

The pretrained reference is essential. Random weights provide minimal signal, and self-reference yields chance performance (confirming pipeline correctness as a sanity check).

\paragraph{Cross-Architecture Reference.}

Can a smaller model serve as reference?

\begin{center}
\begin{tabular}{lc}
\toprule
Reference Architecture & AUC \\
\midrule
Same (GPT-2-XL $\rightarrow$ GPT-2-XL) & 0.895 \\
Cross (DistilGPT-2\textsuperscript{*} $\rightarrow$ GPT-2-XL) & 0.789 \\
\bottomrule
\end{tabular}
\\[2pt]
{\footnotesize \textsuperscript{*}\citet{sanh2019distilbert}}
\end{center}

Same-architecture reference performs substantially better ($-0.106$ AUC for cross-architecture).

\paragraph{Unrelated Domain Reference.}

Does the reference model need to match the target's domain?

\begin{center}
\begin{tabular}{lc}
\toprule
Reference Model & AUC \\
\midrule
Pretrained GPT-2-XL & 0.895 \\
WikiText fine-tuned GPT-2-XL & 0.864 \\
\bottomrule
\end{tabular}
\end{center}

A reference fine-tuned on unrelated domain data (WikiText) still achieves strong performance, only 0.031 AUC below the pretrained baseline. This suggests EZ is robust to reference model choice.

\paragraph{Distillation Reference.}

Following \citet{fu2024practical}, we evaluate a distillation-based reference constructed by prompting the target model and fine-tuning on its generations.

\textbf{Construction:} We prompt the fine-tuned target with up to 10,000 seed texts from related public datasets (News Category Dataset \citep{misra2022news} for AGNews, CNN/DailyMail \citep{hermann2015teaching} for XSum, and Wikipedia for Wikitext, truncated to 16 tokens), generate 112-token completions using nucleus sampling \citep{holtzman2020curious} ($p = 0.9$, temperature $= 0.9$), and fine-tune a fresh copy of the base model on these synthetic texts for 4 epochs.

\begin{center}
\begin{tabular}{lcc}
\toprule
Reference & AUC & Additional Training \\
\midrule
Pretrained (Base) & 0.895 & None \\
Distillation & 0.891 & 4 epochs \\
\bottomrule
\end{tabular}
\end{center}

Distillation provides no improvement over the simpler pretrained reference on this configuration while requiring additional computation. We recommend the pretrained reference as the default choice.

\subsection{Aggregation Function}

We compare alternative ways to aggregate the $\delta$ values at error positions:

\begin{center}
\begin{tabular}{lc}
\toprule
Aggregation & AUC \\
\midrule
Positive fraction & 0.904 \\
Median $\delta$ & 0.899 \\
$P - N$ (difference) & 0.898 \\
$P / N$ (EZ) & 0.895 \\
$\log(P / N)$ & 0.895 \\
Mean $\delta$ & 0.877 \\
\bottomrule
\end{tabular}
\end{center}

Several aggregations achieve similar performance. We select $P/N$ (EZ) for its theoretical grounding (scale invariance, interpretability as odds ratio) despite slightly lower AUC than positive fraction. All methods show high correlation ($r > 0.75$).

\subsection{Error Definition}

We vary the definition of ``error'' from top-1 (model's best prediction is wrong) to top-$K$ (correct token not in top $K$ predictions):

\begin{center}
\begin{tabular}{lcc}
\toprule
Definition & AUC & Error Fraction \\
\midrule
Top-1 & 0.895 & 50.6\% \\
Top-5 & 0.876 & 25.9\% \\
Top-10 & 0.855 & 18.6\% \\
\bottomrule
\end{tabular}
\end{center}

Top-1 is optimal. Stricter definitions (top-5, top-10) reduce the number of positions considered and discard useful signal.

\subsection{Zero-Error Sample Handling}

Sequences where the model predicts every token correctly (zero errors) yield undefined EZ. In our evaluation, 0/1000 members and 0/1000 non-members had zero errors, so all strategies yield identical results. For robustness, we recommend treating such samples as members.

\subsection{N=0 Edge Case}

When all probability movement at error positions is upward ($N=0$, $P>0$), the EZ score $P/N$ is undefined. However, this case represents the strongest possible membership signal—fine-tuning increased probability at every error position. In practice, we assign $\text{EZ} = \infty$ (or a large constant) to such sequences, classifying them as members. This edge case is rare; in our evaluation, 0/1000 members and 0/1000 non-members exhibited $N=0$.

\subsection{Scaling with Training Set Size}
\label{sec:member-scaling}

We evaluate robustness to training set size by scaling from 10k to 250k members on WikiText-103 with Qwen-2.5-1.5B \citep{yang2024qwen25} (LoRA). All methods degrade as the per-sample memorization signal weakens with more training data, but EZ-MIA degrades most gracefully and remains the top method at every scale.

\begin{table}[h]
\centering
\small
\begin{tabular}{lccccc}
\toprule
Members & LOSS & Zlib & MK++ & RefL & EZ-MIA \\
\midrule
10,000  & .583 & .598 & .568 & .684 & \textbf{.870} \\
50,000  & .562 & .573 & .550 & .629 & \textbf{.797} \\
250,000 & .542 & .552 & .533 & .595 & \textbf{.724} \\
\bottomrule
\end{tabular}
\caption{AUC as training set size increases (WikiText-103, Qwen-2.5-1.5B, LoRA). EZ-MIA maintains the largest margin over baselines at all scales.}
\label{tab:member-scaling}
\end{table}

\section{Extended Model and Dataset Evaluation}
\label{sec:extended_eval}

To validate generalizability beyond the primary evaluation, we test on three additional datasets spanning different domains and languages: Enron emails,\footnote{\texttt{SetFit/enron\_spam} on HuggingFace.} PubMed abstracts,\footnote{\texttt{ccdv/pubmed-summarization} on HuggingFace.} and mC4-German.\footnote{\texttt{mc4}, German split, on HuggingFace.} We evaluate four additional models: DistilGPT2 (82M, full fine-tuning), Gemma-3-1B (LoRA), DeepSeek-R1-Distill-Llama-8B (LoRA), and Qwen3-14B (LoRA). SPV-MIA is excluded as its computational requirements (reference model training plus $\sim$42 forward passes per sample) are prohibitive at the 8B and 14B scales. Fine-tuning configurations match those in Table~\ref{tab:finetune_config}.

\begin{table*}[t]
\centering
\small
\resizebox{\textwidth}{!}{%
\begin{tabular}{llcccccccccc}
\toprule
& & \multicolumn{5}{c}{\textbf{AUC}} & \multicolumn{5}{c}{\textbf{TPR@1\%FPR (\%)}} \\
\cmidrule(lr){3-7} \cmidrule(lr){8-12}
\textbf{Dataset} & \textbf{Model} & \textbf{LOSS} & \textbf{Zlib} & \textbf{MK++} & \textbf{RefL} & \textbf{EZ-MIA} & \textbf{LOSS} & \textbf{Zlib} & \textbf{MK++} & \textbf{RefL} & \textbf{EZ-MIA} \\
\midrule
Enron & DistilGPT2 (82M) & .581 & .586 & .587 & .608 & \textbf{.755} & 1.5 & 1.4 & 1.4 & 1.6 & \textbf{1.7} \\
Enron & Gemma-3-1B & .560 & .565 & .554 & .609 & \textbf{.721} & 1.8 & 1.7 & 1.2 & \textbf{1.8} & 1.8 \\
Enron & DeepSeek-8B & .674 & .688 & .668 & .739 & \textbf{.886} & 2.9 & 2.8 & 2.1 & 2.7 & \textbf{3.3} \\
Enron & Qwen3-14B & .645 & .657 & .644 & .734 & \textbf{.871} & 2.4 & 2.4 & 1.9 & 2.4 & \textbf{3.4} \\
\midrule
PubMed & DistilGPT2 (82M) & .644 & .652 & .636 & .773 & \textbf{.902} & 1.7 & 2.1 & 1.3 & 3.3 & \textbf{22.0} \\
PubMed & Gemma-3-1B & .533 & .536 & .533 & .653 & \textbf{.701} & 1.4 & 1.6 & 1.1 & 3.2 & \textbf{4.6} \\
PubMed & DeepSeek-8B & .719 & .731 & .707 & .917 & \textbf{.952} & 3.1 & 5.7 & 1.9 & 28.8 & \textbf{46.4} \\
PubMed & Qwen3-14B & .552 & .556 & .553 & .699 & \textbf{.721} & 1.3 & 1.7 & 1.2 & 5.9 & \textbf{6.5} \\
\midrule
mC4-German & DistilGPT2 (82M) & .594 & .596 & .590 & .669 & \textbf{.765} & 1.5 & 1.8 & 1.8 & 3.8 & \textbf{5.7} \\
mC4-German & Gemma-3-1B & .528 & .528 & .521 & .685 & \textbf{.753} & 1.0 & 0.9 & 1.0 & 4.1 & \textbf{7.1} \\
mC4-German & DeepSeek-8B & .597 & .595 & .569 & .775 & \textbf{.813} & 1.4 & 1.5 & 1.3 & 7.1 & \textbf{9.3} \\
mC4-German & Qwen3-14B & .559 & .559 & .547 & .738 & \textbf{.792} & 1.4 & 1.4 & 1.2 & 3.5 & \textbf{7.4} \\
\bottomrule
\end{tabular}
}
\caption{Extended evaluation across additional models and datasets. DistilGPT2 uses full fine-tuning; all other models use LoRA ($r{=}16$, $\alpha{=}32$). Bold indicates best per row per metric. EZ-MIA achieves the highest AUC in all 12 configurations.}
\label{tab:extended_eval}
\end{table*}

\end{document}